\documentclass[10pt,reqno]{amsart}
\usepackage{indentfirst,csquotes}
\usepackage{amsmath,amsfonts,bm}

\def\eqref#1{equation~\ref{#1}}
\def\1{\bm{1}}

\DeclareMathAlphabet{\mathsfit}{\encodingdefault}{\sfdefault}{m}{sl}
\SetMathAlphabet{\mathsfit}{bold}{\encodingdefault}{\sfdefault}{bx}{n}

\usepackage{hyperref}
\usepackage{url}
\usepackage{enumitem}
\usepackage{caption}
\usepackage{microtype}
\usepackage{graphicx}
\usepackage{subfigure}
\title{Federated Conditional Conformal Prediction via Generative Models}

\author{Rui Xu$^1$, Xingyuan Chen$^1$, Wenxing Huang$^2$, \\Minxuan Huang$^2$, Yun Xie$^3$, Weiyan Chen$^2$, Sihong Xie$^1$}

\thanks{$^1$ Information Hub, Hong Kong University of Science and Technology (Guangzhou)}
\thanks{$^2$ Critical Care Medicine, The Second Affiliated Hospital of Guangzhou Medical University}
\thanks{$^3$ Critical Care Medicine, Shanghai General Hospital, Shanghai Jiao Tong University School of Medicine}

\begin{document}

\maketitle
\begin{abstract}
Conformal Prediction (CP) provides distribution-free uncertainty quantification by constructing prediction sets that guarantee coverage of the true labels. This reliability makes CP valuable for high-stakes federated learning scenarios such as multi-center healthcare. However, standard CP assumes i.i.d. data, which is violated in federated settings where client distributions differ substantially. Existing federated CP methods address this by maintaining marginal coverage on each client, but such guarantees often fail to reflect input-conditional uncertainty. In this work, we propose Federated Conditional Conformal Prediction (Fed-CCP) via generative models, which aims for conditional coverage that adapts to local data heterogeneity. Fed-CCP leverages generative models, such as normalizing flows or diffusion models, to approximate conditional data distributions without requiring the sharing of raw data. This enables each client to locally calibrate conformal scores that reflect its unique uncertainty, while preserving global consistency through federated aggregation. Experiments on real datasets demonstrate that Fed-CCP achieves more adaptive prediction sets.
\end{abstract}

\section{Introduction}
Uncertainty in model predictions continues to limit the safe adoption of artificial intelligence in critical domains such as finance~\cite{ryu2020sustainable}, healthcare~\cite{lambert2024trustworthy}, and autonomous driving~\cite{fingscheidt2022deep}. To quantify this uncertainty, a range of approaches have been proposed~\cite{guo2017calibrationmodernneuralnetworks,jospin2022hands,ovadia2019can}, among which Conformal Prediction (CP) stands out for providing distribution-free coverage guarantees. Rather than producing a single point value with estimated confidence, CP constructs a prediction set that contains the true label with a user-specified confidence level~\cite{vovk2005algorithmic,shafer2007tutorialconformalprediction,angelopoulos2021gentle}. In the regression setting~\cite{lei2017distributionfreepredictiveinferenceregression}, given a predictive model $h$, CP defines a score function $s(X,Y)=|h(X)-Y|$ that measures residuals on a calibration dataset $\{(X_i,Y_i)\}_{i=1}^n$. 
Let $\tau$ denote the $\lceil(1-\alpha)(n+1)\rceil/n$ quantile of these calibration scores. 
The standard (marginal) conformal prediction set is then given by $C_\mathrm{M}(X_{n+1}) = \{y: s(X_{n+1},y) \le \tau \}$.
When the calibration and test samples are exchangeable such that the calibration and test distributions coincide, i.e., $P_{XY} = Q_{XY}$, the prediction set $C_\mathrm{M}(X_{n+1})$ satisfies the \textbf{marginal coverage guarantee}: $\Pr\!\big(Y_{n+1} \in C_\mathrm{M}(X_{n+1})\big) \ge 1 - \alpha$.

Federated Learning (FL) enables multiple institutions or devices to collaboratively train machine learning models without directly sharing raw data~\cite{wen2023survey}. 
This paradigm is particularly valuable in privacy-sensitive and data-siloed domains where data governance or legal constraints prohibit centralized data aggregation. 
Despite its promise, FL introduces substantial challenges for CP, because client datasets are often \emph{non-exchangeable}, leading to invalid coverage guarantees. To address this challenge, FCP~\cite{lu2021distribution,lu2023federated} and FedCP-QQ~\cite{humbert2023one} aim for a coverage guarantee when the test sample is drawn from a known mixture of client distributions, and ~\cite{liu2024multi} further study the situation of missing outcomes in calibration data.
In contrast, DP-FedCP~\cite{plassier2023conformal} considers a more challenging setting where test samples originate from a single client distribution, and heterogeneity among clients arises primarily from label shift ($P_Y \neq Q_Y$). Moreover,~\cite{wen2025distributedconformalpredictionmessage} explore a more realistic setting in which clients communicate only with their neighbors over an arbitrary graph topology, without requiring direct connections to the central server.

However, marginal coverage leads to prediction intervals of uniform width that fail to reflect instance-specific uncertainty. 
This lack of adaptiveness motivates the development of \textbf{conditional} or \textbf{adaptive} CP, which seeks to ensure $
\Pr\!\big(Y_{n+1} \in C_\mathrm{A}(X_{n+1}) \mid X_{n+1} = x\big) \ge 1 - \alpha, \quad \forall x \in \mathcal{X}$,
thereby achieving \textbf{conditional coverage guarantee} that adjusts prediction sets according to the local difficulty or uncertainty of each input~\cite{papadopoulos2011regression,vovk2012conditional}. Prior federated CP methods typically maintain marginal coverage by comparing calibration and test conformal score distributions, but they do not investigate how to achieve conditional coverage in federated settings.

In this work, we propose Federated Conditional Conformal Prediction (Fed-CCP) to provide instance-specific prediction sets for each client with two key contributions.
\begin{enumerate}[topsep=0pt, itemsep=0pt, leftmargin=15pt]
    \item \textbf{Conditional coverage guarantee by transformable prediction set.} On the global server, we define a simple Gaussian distribution and leverage a generative model $f$, such as a diffusion model or a normalizing flow, to construct a mapping between each client's data and the Gaussian distribution. 
    This allows prediction sets defined in the Gaussian space to be transformed back to the original data space of each client. By doing so, we can approximate conditional coverage guarantees in the federated setting, adapting prediction sets to local data heterogeneity. 

    \item \textbf{Privacy preservation.} Training of the generative model relies solely on gradient exchanges between the server and clients, ensuring that raw client data never leaves local devices. This design not only safeguards sensitive information but also enables the global generative model to capture the underlying data distribution across clients without direct access to private datasets. Furthermore, the base predictive model $h$ is trained using samples drawn from the simple Gaussian distribution, thereby eliminating any need to access client data.
\end{enumerate}

We proved the effectiveness of Fed-CCP on real-world datasets, demonstrating that it produces tighter and more adaptive prediction sets while achieving more uniform conditional coverage across heterogeneous clients compared to existing methods.

\section{Background}
Let $X \in \mathcal{X} \subseteq \mathbb{R}^d$ and $Y \in \mathcal{Y} \subseteq \mathbb{R}$ denote the input and output random variables, respectively.  
Given a regression model $h: \mathcal{X} \rightarrow \mathcal{Y}$, we define a score function $s: \mathcal{X} \times \mathcal{Y} \rightarrow \mathcal{V} \subseteq \mathbb{R}$ that measures how well a pair $(x,y)$ aligns with the model’s prediction.  
The corresponding random variable $V = s(X,Y)$, referred to as the \emph{conformal score}, commonly takes the form of the absolute residual $|h(X)-Y|$.  

Suppose we have calibration samples $\{(X_i, Y_i)\}_{i=1}^n$ drawn from a distribution $P_{XY}$.  
The split conformal prediction procedure~\cite{papadopoulos2002inductive} computes the calibration scores $\{V_i = s(X_i, Y_i)\}_{i=1}^n$.  
For a new test point $(X_{n+1}, Y_{n+1})$ from distribution $Q_{XY}$, the conformal prediction interval is defined as  $C_\mathrm{M}(X_{n+1}) = \{y \in \mathcal{Y} : s(X_{n+1}, y) \le \tau\}$,
where $\tau$ is chosen as the ${\lceil(1-\alpha)(n+1)\rceil}/{n}$ empirical quantile of the calibration scores. 
When the calibration and test distributions coincide ($P_{XY} = Q_{XY}$), this construction satisfies the \textbf{marginal coverage guarantee}:
\[
\Pr\!\big(Y_{n+1} \in C_\mathrm{M}(X_{n+1})\big) \ge 1 - \alpha.
\]
However, since $\tau$ is a fixed quantile independent of $X_{n+1}$, the resulting prediction intervals have uniform width across all inputs.  
This uniformity leads to miscalibrated uncertainty estimates, which are overly wide for simple samples and too narrow for difficult ones~\cite{angelopoulos2022conformal}.  
To overcome this limitation, adaptive conformal prediction introduces a data-dependent threshold $\tau(x)$, yielding prediction sets
$C_\mathrm{A}(X_{n+1}) = \{y \in \mathcal{Y} : s(X_{n+1}, y) \le \tau(x)\}$,
where $\tau(x)$ represents the ${\lceil(1-\alpha)(n_x+1)\rceil}/{n_x}$ quantile of $\{V_i : X_i = x\}$ and $n_x$ is the number of calibration samples whose feature are $x$.  
This construction targets the \textbf{conditional coverage guarantee}~\cite{vovk2012conditional}:
\begin{equation} \label{eq: conditional guarantee}
\Pr\!\big(Y_{n+1} \in C_\mathrm{A}(X_{n+1}) \mid X_{n+1}=x\big) \ge 1 - \alpha, \quad \forall x \in \mathcal{X}.
\end{equation}
The conditional guarantee is unattainable with limited calibration data without extra assumptions, motivating extensive work on its practical approximations~\cite{bostrom2021mondrian,romano2019conformalized,lin2021locally, colombo2024normalizing}.

\section{Method}
\subsection{Generative models for distribution transformation}

To mitigate distributional heterogeneity across clients, we employ generative models to learn an invertible transformation between a client data distribution $Q_{XY}$ and a reference calibration distribution $P_{XY}$, both defined over the same space $\mathcal{X}\times\mathcal{Y}$. 
Let $f$ denote the \textbf{forward mapping} that transforms samples from $Q_{XY}$ toward $P_{XY}$, and let $g$ denote its \textbf{inverse mapping} that reconstructs client data from the reference space. Thereby, we can denote $g=f^{-1}$.
This bidirectional relationship enables prediction sets defined under the calibration distribution to be consistently transformed back to the client’s local data space.

\paragraph{Normalizing Flows.}
% Normalizing flows (NFs)~\cite{rezende2016variationalinferencenormalizingflows,papamakarios2021normalizing} are a family of generative models that learn an invertible transformation between two continuous random variables with tractable probability densities. 
% Let $q_{XY}(x,y)$ denote the probability density of $Q_{XY}$ at a real-valued sample $(x,y)$. Besides, let $f_\#Q_{XY}$ be the push forward distribution from $Q_{XY}$ by $f$ with density denoted by $f_\#q_{XY}$.
% The change-of-variables formula allows the density of $f_\#Q_{XY}$ to be expressed in terms of  $q_{XY}$ as
% \begin{equation}\label{eq:nf_density}
%     f_\#q_{XY}(f(x,y)) = q_{XY}\big(x,y\big)\left|\det \frac{\partial f(x,y)}{\partial (x,y)}\right|^{-1},
% \end{equation}
% where $\frac{\partial f(x,y)}{\partial (x,y)}$ denotes the Jacobian matrix of $f$ with respect to $(x,y)$.

Normalizing flows (NFs)~\cite{rezende2016variationalinferencenormalizingflows,papamakarios2021normalizing} are a family of generative models that learn an invertible transformation between two continuous random variables with tractable probability densities.  
Let $q_{XY}(x,y)$ denote the probability density of the source distribution $Q_{XY}$ at a real-valued sample $(x,y)$.  
Let $f_\# Q_{XY}$ denote the pushforward distribution of $Q_{XY}$ under the bijective transformation $f$, whose density is written as $f_\# q_{XY}$.  
According to the change-of-variables formula, the density of $f_\# Q_{XY}$ at the forward sample $f(x,y)$ can be expressed in terms of $q_{XY}$ as
\begin{equation}\label{eq:nf_density}
    f_\# q_{XY}\big(f(x,y)\big)
    = q_{XY}(x,y)
      \left|\det \frac{\partial f(x,y)}{\partial (x,y)}\right|^{-1},
\end{equation}
where $\frac{\partial f(x,y)}{\partial (x,y)}$ denotes the Jacobian matrix of $f$ with respect to its input. 

Given this relationship, the parameters of $f$ (and its inverse $g=f^{-1}$) can be learned by minimizing the divergence between the induced density $f_\# q_{XY}$ and the target distribution density $p_{XY}$.  
In practice, this is commonly formulated as the following optimization problem:
\begin{equation*}
    \min_\theta D_{\mathrm{KL}}\Big(f_\theta{}_\# q_{XY} \;\|\; p_{XY}\Big)
    = \min_\theta - \mathbb{E}_{(x,y)\sim Q_{XY}} \Big[ \log p_{XY}\big(f_\theta(x,y)\big) 
    + \log \Big|\det \frac{\partial f_\theta(x,y)}{\partial (x,y)}\Big|^{-1} \Big],
\end{equation*}
where $\theta$ denotes the parameters of the flow model and $D_{\mathrm{KL}}(\cdot\|\cdot)$ is the Kullback--Leibler divergence.  
Minimizing this objective aligns the transformed client distribution with the reference calibration distribution while preserving invertibility, allowing the inverse mapping $g_\theta = f_\theta^{-1}$.

\paragraph{Diffusion Models.}
Diffusion models~\cite{ho2020denoisingdiffusionprobabilisticmodels,song2021scorebasedgenerativemodelingstochastic} achieve a distributional transformation through a stochastic forward–reverse process.  
In the forward process $f$, $(X,Y)\sim Q_{XY}$ is gradually perturbed toward the reference calibration distribution $P_{XY}$ by adding Gaussian noise over a continuous time variable $t \in [0, T]$, where $t$ controls the noise level and $T$ is the final diffusion time.  
This process is typically described by the stochastic differential equation (SDE):
\begin{equation}
    \mathrm{d}(X_t,Y_t) = -\frac{1}{2}\beta_t(X_t,Y_t)\,\mathrm{d}t + \sqrt{\beta_t}\,\mathrm{d}W_t,
\end{equation}
where $\beta_t$ is the time-dependent diffusion coefficient and $W_t$ denotes standard Brownian motion.  

The reverse process, parameterized by a neural network, learns to progressively denoise samples starting from $t=T$ back to $t=0$, effectively approximating the inverse mapping $g \approx f^{-1}$.  
After training, $g$ can transform samples from the reference distribution $P_{XY}$ back to the original data distribution $Q_{XY}$, providing a flexible generative model for downstream tasks.

Both normalizing flows and diffusion models enable invertible transformations between $Q_{XY}$ and $P_{XY}$ within the same feature–label space. 
This property allows conformal prediction sets to be constructed under a well-behaved reference calibration distribution and then consistently projected back to heterogeneous client domains through the learned mappings $(f,g)$, forming the foundation of our Federated Conditional Conformal Prediction framework.

\subsection{Transformable prediction set}
Let $f_\theta: \mathcal{X}\times\mathcal{Y} \rightarrow \mathcal{X}\times\mathcal{Y}$ denote a learned invertible transformation that maps the client distribution $Q_{XY}$ to a reference calibration distribution $P_{XY}$. 
Then, for any test instance $(X_{n+1}, Y_{n+1})$ drawn from the client distribution, we have
\begin{equation}
    f_\theta(X_{n+1}, Y_{n+1}) = (\tilde{X}_{n+1}, \tilde{Y}_{n+1}) \sim P_{XY}, 
    \quad \forall (X_{n+1}, Y_{n+1}) \sim Q_{XY}.
\end{equation}
Since $(\tilde{X}_{n+1}, \tilde{Y}_{n+1})$ and the calibration samples $\{(X_i,Y_i)\}_{i=1}^n$ are exchangeable under $P_{XY}$, a conformal prediction set constructed using the calibration samples,
\begin{equation}
    C_\mathrm{A}(\tilde{X}_{n+1}) = \big\{ \tilde{y}\in \mathcal{Y} : s(\tilde{X}_{n+1}, \tilde{y}) \le \tau(\tilde{X}_{n+1}) \big\},
\end{equation}
achieves the conditional coverage guarantee:
\begin{equation}
    \Pr\big(\tilde{Y}_{n+1} \in C_\mathrm{A}(\tilde{X}_{n+1}) \mid \tilde{X}_{n+1}=\tilde{x}\big) 
    \ge 1 - \alpha, ~ \forall \tilde{x}\in\mathcal{X}.
\end{equation}

By leveraging the bijectivity of $f_\theta$, we can project this prediction set back to the original client distribution $Q_{XY}$ to obtain the \textbf{transformable prediction set}:
\begin{equation}
    C_\mathrm{Trans}(X_{n+1}) = 
    \big\{ y \in \mathcal{Y} : f_\theta(X_{n+1}, y) = (\tilde{X}_{n+1}, \tilde{y}),~ \tilde{y}\in C_\mathrm{A}(\tilde{X}_{n+1}) \big\}.
\end{equation}
Equivalently, using the inverse mapping $g_\theta = f_\theta^{-1}$, this can be written as
\begin{equation}
    C_\mathrm{Trans}(X_{n+1}) = 
    \big\{ y \in \mathcal{Y} : \tilde{y} \in C_\mathrm{A}(\tilde{X}_{n+1}),~ g_\theta(\tilde{X}_{n+1}, \tilde{y})=({X}_{n+1}, {y}) \big\}.
\end{equation}
Because $f_\theta$ is bijective, there exists a one-to-one correspondence between instances in $P_{XY}$ and $Q_{XY}$. 
Therefore, the coverage guarantee is preserved under the transformation:
\begin{equation}
    \Pr\big(Y_{n+1} \in C_\mathrm{Trans}(X_{n+1}) \mid X_{n+1}=x\big)
    \ge 1 - \alpha, ~ \forall x \in \mathcal{X}.
\end{equation}
This formulation enables conditional conformal prediction on the shifted client distribution by performing coverage estimation under the reference calibration distribution $P_{XY}$, while preserving valid coverage under the client distribution $Q_{XY}$ through the bijective transformation $g_\theta =f_\theta^{-1}$.

\subsection{Federated learning with multi-clients}
In realistic federated settings, data are distributed across multiple clients, each following a distinct local distribution. 
Let there be $K$ clients, with the data distribution of client $k$ denoted by $Q^{(k)}_{XY}$ for $k = 1, \dots, K$. 
These local distributions may differ in both input and output spaces due to data heterogeneity, domain shifts, or varying data collection processes.

To achieve a unified generative mapping across clients while preserving client-specific characteristics, 
we introduce a conditional generative model $f_\theta(x, y; \eta)$, where $\eta$ is a \emph{client conditioner} that modulates the transformation according to the client identity or domain context. 
Specifically, for client $k$, the conditioner is sampled from a client-specific Gaussian distribution:
\begin{equation}
    \eta^{(k)} \sim \mathcal{N}\big(\mu^{(k)}, \sigma{(k)^2}\big),
\end{equation}
where $\mu^{(k)}$ and $\sigma^{(k)}$ represent the mean and standard deviation associated with client $k$. 
This stochastic conditioner $\eta^{(k)}$ can be interpreted as a representation of client-level variation, allowing $f_\theta$ to flexibly adapt the mapping from $Q^{(k)}_{XY}$ to the reference calibration distribution $P_{XY}$ and maintaining a shared parameterization across all clients.
\subsection{Practical Implementation of Fed-CCP}
In practical applications, only finite samples are available. 
We denote by $S_P$ and $S_{Q^{(k)}}$ the sample sets drawn from the reference calibration distribution $P_{XY}$ and the client distribution $Q_{XY}^{(k)}$, respectively. 
The reference distribution $P_{XY}$ is specified as a multivariate normal with a diagonal covariance matrix, where all diagonal entries are positive and all off-diagonal entries are zero. Moreover, each time a sample is drawn from ${Q^{(k)}_{XY}}$, a corresponding conditioner is also sampled from $\mathcal{N}(\mu^{(k)}, \sigma{(k)^2})$.   

The Fed-CCP procedure can be decomposed into two principal steps:

\begin{enumerate}[topsep=0pt, itemsep=0pt, leftmargin=15pt]
    \item \textbf{Construction of adaptive prediction sets $C_\mathrm{A}$.}  
    Using only the reference samples $S_P$, a base predictive model $h$ is trained via Conformalized Quantile Regression (CQR)~\cite{romano2019conformalized} to approximate conditional coverage.  
    This step does not require access to client data, thereby preserving privacy while establishing a predictive model in the reference distribution.

    \item \textbf{Training the generative model $f_\theta$.}  
    A bijective generative model $f_\theta$ is trained to map client samples to the reference calibration distribution:
    \[
        f_\theta: (X,Y;\eta) \mapsto (\tilde{X},\tilde{Y}) \sim P_{XY}.
    \]  
    In this step, client datasets $\{S_{Q^{(k)}}\}_{k=1}^K$ are used.   
    Only gradient information is communicated to the server, ensuring that raw client data remains private while the global generative model captures the shared structure across heterogeneous client distributions.
\end{enumerate}

\section{Experiment}
\subsection{Experimental Setup}
We implement Fed-CCP using the \texttt{normflows} library~\cite{Stimper2023}. 
For comparison, we include Conformalized Quantile Regression (CQR) without the generative model component to demonstrate that CQR alone fails to maintain valid conditional coverage under federated distribution shifts. 
Additionally, we evaluate Fed-CCP without the conditioner $\eta$ in the generative model, showing that the absence of client-specific conditioning significantly undermines performance. 

\textbf{Datasets.} Multiple datasets are employed in our experiments to evaluate Fed-CCP across diverse real-world federated learning scenarios. In the healthcare domain, we aim to ensure unbiased and reliable predictions for patients across different hospitals. We leverage data from MIMIC-IV~\cite{johnson2023mimic}, eICU~\cite{pollard2018eicu}, and two collaborating hospitals, representing heterogeneous medical centers with varying patient populations and clinical practices. This setup reflects the common challenge of distributional heterogeneity in multi-center healthcare applications, where direct data sharing is often restricted due to privacy regulations. In the Internet of Things (IoT) setting, we use the Intel Berkeley Research Lab Sensor Data~\cite{madden2003intel} to simulate a federated sensor network, where each client corresponds to a different group of sensors deployed across spatially separated environments. This scenario captures device-level data drift caused by local environmental differences. For the insurance domain, we utilize the French Motor Claims dataset~\cite{dutang2020package}, which contains 677,991 third-party motor liability policies. We use policyholder and vehicle attributes to predict the Bonus-Malus coefficient while ensuring privacy protection. Each client represents a distinct geographic region (e.g., city or province), allowing us to investigate Fed-CCP’s robustness under geographically induced heterogeneity.
In the traffic forecasting domain, we consider a realistic federated scenario where data originates from heterogeneous sources such as government traffic departments, private companies, and individual users. We use Seattle-Loop~\cite{cui2019traffic}, PEMSD4, and PEMSD8~\cite{bai2020adaptivegraphconvolutionalrecurrent} datasets to perform federated traffic speed prediction, where each data provider corresponds to a distinct client.
Finally, for epidemic modeling, we employ datasets from US-Regions, US-States, and Japan-Prefectures~\cite{deng2020cola} to conduct federated epidemic spread prediction. Each client corresponds to a specific administrative region, enabling the study of Fed-CCP’s ability to model and adapt to spatially distributed epidemiological dynamics under strict data locality constraints.

\textbf{Metrics.} Since the validity of conditional coverage guarantees cannot be directly measured using finite samples, we adopt two surrogate metrics for evaluation.
First, we assess the marginal coverage guarantee, which serves as a necessary condition for conditional validity—if conditional coverage holds, marginal coverage must also be satisfied.
Second, we evaluate the average prediction set size, as smaller sets indicate higher adaptivity to instance-specific uncertainty and thus better approximation of conditional coverage.
Together, these two metrics provide a practical and interpretable assessment of how well the proposed method captures conditional reliability under limited data.
\begin{figure*}[h]
\centering
\captionsetup{singlelinecheck = false, justification=justified}
  \includegraphics[scale=0.4]{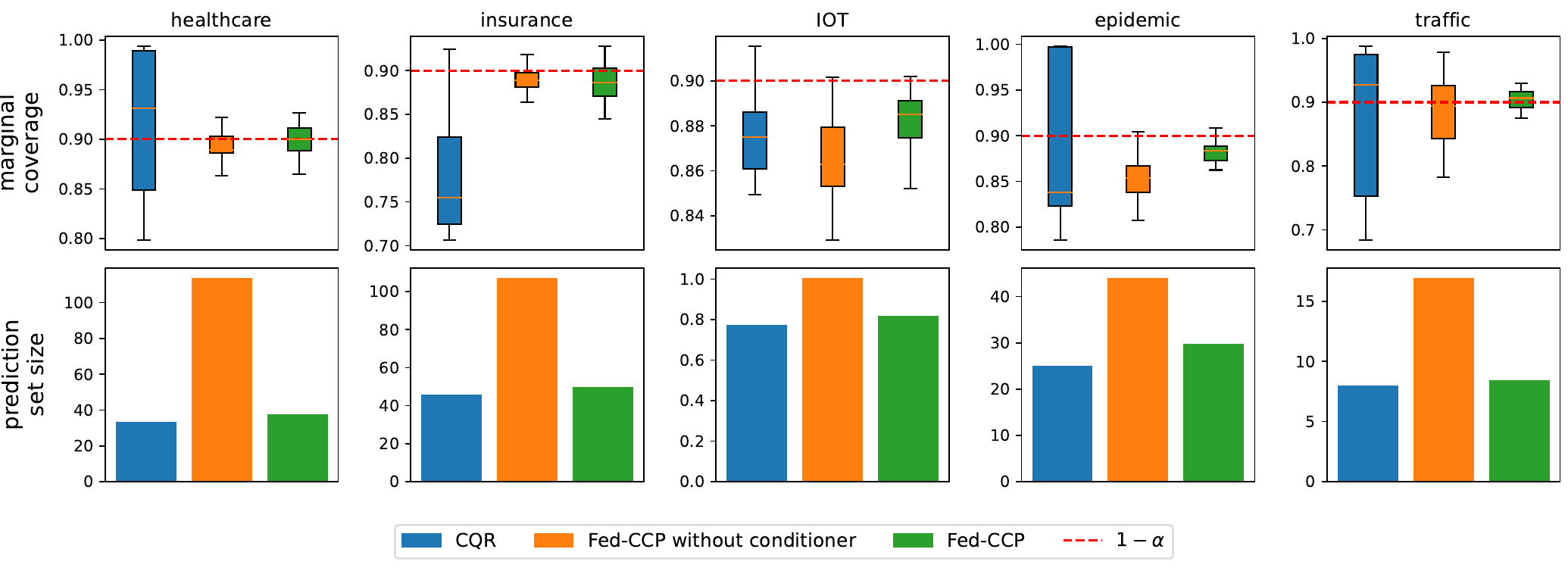}
  \caption{\footnotesize Comparison of CQR, Fed-CCP (w/o conditioner), and Fed-CCP. The top row shows marginal coverage (red dashed line: target 0.9), and the bottom row shows average prediction set size. Fed-CCP attains near-nominal coverage with smaller sets, indicating better adaptiveness and efficiency.}
  \label{fig: results} 
  % \vspace{-10pt}
\end{figure*}
\subsection{Result}
Figure~\ref{fig: results} summarizes the performance of different methods across five domains: healthcare, insurance, IoT, epidemic forecasting, and traffic prediction.
The upper row reports the marginal coverage, while the lower row shows the average prediction set size. The red dashed line indicates the nominal confidence level $1-\alpha=0.9$. Across all tasks, Fed-CCP consistently achieves marginal coverage close to the target level while maintaining the relatively small prediction set size among all methods.
This indicates that Fed-CCP not only preserves the coverage validity under heterogeneous client distributions but also adapts effectively to instance-specific uncertainty. By contrast, CQR without the generative model exhibits large variability in coverage and often undercovers in federated settings, reflecting its sensitivity to distributional shifts.
Furthermore, Fed-CCP without the conditioner fails to capture client-specific characteristics, resulting in inflated prediction sets.
These results confirm that the incorporation of the conditioner $\eta$ is crucial for modeling cross-client heterogeneity and that the bijective generative alignment in Fed-CCP effectively harmonizes client distributions to maintain reliable and adaptive uncertainty quantification.

\section{Conclusion}
In this work, we introduced Federated Conditional Conformal Prediction (Fed-CCP), a framework for producing instance-specific prediction sets in federated learning environments. By leveraging a generative mapping between client data and a simple Gaussian calibration distribution, Fed-CCP transforms prediction sets defined in the Gaussian probability space back to each client’s original data space, enabling adaptive prediction intervals that approximate conditional coverage guarantees. Importantly, our approach preserves privacy by relying solely on gradient exchanges, without requiring access to raw client data.
\bibliographystyle{ieeetr}
\bibliography{reference}

\end{document}